\pdfoutput=1

\documentclass[11pt]{article}

\usepackage{acl}

\usepackage{times}
\usepackage{latexsym}

\usepackage[T1]{fontenc}

\usepackage[utf8]{inputenc}

\usepackage{microtype}

\usepackage{inconsolata}
\usepackage{amssymb}
\usepackage{amsthm}
\usepackage{amsmath}
\usepackage{multirow}
\usepackage{graphicx}
\usepackage{enumitem}
\usepackage{bm}
\usepackage{url}
\usepackage{hyperref}
\usepackage{todonotes}
\usepackage{subfiles}

\usepackage{caption}
\usepackage{subcaption}

\title{STT4SG-350: A Speech Corpus for All Swiss German Dialect Regions}

\author{
Michel Plüss\textsuperscript{1}, Jan Deriu\textsuperscript{2}, Yanick Schraner\textsuperscript{1}, Claudio Paonessa\textsuperscript{1}, \\ 
{\bf Julia Hartmann\textsuperscript{1},} {\bf Larissa Schmidt\textsuperscript{3},}
{\bf Christian Scheller\textsuperscript{1}, }{\bf Manuela Hürlimann\textsuperscript{2}, }
\\{\bf Tanja Samardžić\textsuperscript{3}, }{\bf Manfred Vogel\textsuperscript{1}, }{\bf Mark Cieliebak\textsuperscript{2}} \\
\textsuperscript{1}University of Applied Sciences and Arts Northwestern Switzerland, Windisch \\
\textsuperscript{2}Zurich University of Applied Sciences, Winterthur \\
\textsuperscript{3}University of Zurich, Zurich \\
deri@zhaw.ch, yanick.schraner@fhnw.ch, claudio.paonessa@fhnw.ch, \\ julia.hartmann@fhnw.ch, hueu@zhaw.ch, \\
tanja.samardzic@uzh.ch, manfred.vogel@fhnw.ch, ciel@zhaw.ch
}

\begin{document}

\maketitle

\begin{abstract}

We present STT4SG-350 (Speech-to-Text for Swiss German), a corpus of Swiss German speech, annotated with Standard German text at the sentence level. The data is collected using a web app in which the speakers are shown Standard German sentences, which they translate to Swiss German and record. We make the corpus publicly available. It contains 343 hours of speech from all dialect regions and is the largest public speech corpus for Swiss German to date. Application areas include automatic speech recognition (ASR), text-to-speech, dialect identification, and speaker recognition. Dialect information, age group, and gender of the 316 speakers are provided. Genders are equally represented and the corpus includes speakers of all ages. Roughly the same amount of speech is provided per dialect region, which makes the corpus ideally suited for experiments with speech technology for different dialects. We provide training, validation, and test splits of the data. The test set consists of the same spoken sentences for each dialect region and allows a fair evaluation of the quality of speech technologies in different dialects. We train an ASR model on the training set and achieve an average BLEU score of $74.7$ on the test set. The model beats the best published BLEU scores on 2 other Swiss German ASR test sets, demonstrating the quality of the corpus.

\end{abstract}

\section{Introduction}
\label{sec.intro}
We present STT4SG-350, a corpus of Swiss German speech, annotated with Standard German text at the sentence level. The corpus represents all Swiss German dialect regions and contains 343 hours of speech.

Swiss German is a family of German dialects spoken by around 5 million people in Switzerland. It differs from Standard German regarding phonology, vocabulary, morphology, and syntax. There are significant differences among the Swiss German dialects as well, particularly regarding phonology and vocabulary. Swiss German is primarily a spoken language. It is also used in writing, but mainly in informal text messages. In most other contexts, including formal letters, laws, and newspapers, Standard German is used instead. One important reason for this is Swiss German's lack of a standardized orthography.

The diversity among dialects, exacerbated by the lack of a standardized orthography, leads to a large number of written variants for each word. This, together with the small amount of text resources compared to Standard German, makes automated processing of Swiss German text challenging.

STT4SG-350 is, to the best of our knowledge, the largest public speech corpus for Swiss German. While the primary use case is automatic speech recognition (ASR), it is also a useful resource for text-to-speech (TTS), dialect identification, and speaker recognition. By providing roughly the same amount of data per dialect region, irrespective of its population size, the corpus contributes to improving speech technology for underrepresented dialects. In addition, the test set, which contains the same spoken sentences in each dialect, allows a fair evaluation of the quality of speech technologies in different dialects. Furthermore, it contributes to more inclusive speech technology by keeping a balanced gender ratio and featuring speakers of all ages.


\section{Related Work}
\label{sec.rel}
The SDS-200 corpus~\cite{pluss-etal-2022-sds} contains 200 hours of speech by around 4,000 speakers with Standard German transcripts. The recordings cover a large part of the Swiss German dialect landscape. The number of recordings per speaker follows a long-tail distribution. For example, the top 3 speakers account for 23\% of recordings. The Swiss Parliaments Corpus or SPC~\cite{pluss-etal-2021-spc} contains 299 hours of speech in the Bernese dialect. The text is Standard German, taken from parliament minutes, and is not a fully accurate transcription. Text and audio are automatically aligned. The SwissDial corpus~\cite{dogan-etal-2021-swissdial} contains 26 hours of studio-quality recordings by 8 speakers, each speaking a different dialect, with both Standard German and Swiss German transcripts. The Radio Rottu Oberwallis corpus~\cite{garner-etal-2014-rottu} contains 8 hours of speech transcribed in Swiss German, of which 2 are also transcribed in Standard German. The ArchiMob corpus~\cite{samardzic-etal-2016-archimob} contains 69 hours of speech with Swiss German transcripts.

For Swiss German ASR, the desired output text language is Standard German for the vast majority of use cases. Tackling speech-to-text translation with an end-to-end approach is feasible as shown by \citet{weiss-et-al-2017-speech-translation}. Applying a similar approach to Swiss German ASR and therefore avoiding Swiss German text and its challenges altogether lead to promising results in recent years, see~\cite{pluss-etal-2022-shared-task, khosravani-etal-2021-st, pluss-etal-2022-sds, pluss-etal-2021-spc}.

\citet{dogan-etal-2021-swissdial} experiment with TTS for Swiss German. Their models achieve a 5-scale mean opinion score of 2.9 to 4.1. Importantly, their approach requires Swiss German input text.

\section{Data Collection}
\label{sec.coll}

Data for STT4SG-350 was collected in two phases: 1) the test set with 76 participants from December 2021 until March 2022, and 2) the train and validation sets with 240 participants from May until November 2022.

\subsection{Recording}

Speech was recorded using a web app based on the code\footnote{MPL-2.0 license} by \citet{pluss-etal-2022-sds}. Recordings are made sentence by sentence. The app displays a Standard German sentence, which the participant is asked to translate to Swiss German and speak aloud. A screenshot of the recording functionality can be found in Appendix~\ref{sec.app.screen}. The goal of the translation step is to get a correct, natural-sounding Swiss German sentence in the participant's dialect. We display a popup with examples before the first recording to explain this to participants. We also display a short explanation below the sentence to be recorded. We manually validated the correctness of at least 10 randomly sampled recordings per participant at collection time. In contrast to \citet{pluss-etal-2022-sds}, for phase 2, we recorded 44.1 kHz lossless FLAC audio rather than 32 kHz lossy MP3 audio. The recording quality depends on the microphones used by participants, which range from studio microphones to headsets and laptop microphones. Depending on the microphone, mouse clicks can be audible in recordings.

\subsection{Dialect Regions}


For this work, we divided the Swiss German dialect continuum into 7 dialect regions, listed in Table~\ref{tbl.corpus-reg}, based on the clustering method by \citet{scherrer-stoeckle-2016-dialectometric}\footnote{Population statistics from \url{https://www.bfs.admin.ch}}. The cluster analysis was carried out on 350 phonological, lexical, morphological, and syntactic phenomena. We slightly adjusted the resulting clusters to match the dialect regions commonly used in public discourse more closely. The goal of these adjustments was to make it more intuitive for participants to choose their dialect region. The borders are intentionally fuzzy to give participants the freedom to choose the region that fits their dialect best.

\subsection{Sentence Selection}

Sentences were randomly selected from Swiss newspapers and from parliament minutes of 2 Swiss parliaments. Sentence filtering for newspapers follows \citet{pluss-etal-2022-sds}. The goal of the filtering is to limit sentence complexity to reduce errors in the translation task. For example, only sentences of 5 to 12 words are kept. The newspaper sentences cover a broad range of topics, including culture, finance, science, sports, and technology. They also cover content and named entities particularly relevant for Switzerland. Parliament sentences are not filtered. They bring additional diversity to the corpus with longer sentences on average and a distinct vocabulary. For the test set, 3,515 sentences were selected (67\%  newspapers, and 33\% parliaments). To allow a fair comparison among the dialects, each sentence was recorded in each of the 7 dialects. For the training and validation data, 94\% news and 6\% parliament sentences were selected, and we dropped the requirement to record each sentence in all dialect regions to increase vocabulary and phrase diversity.


\subsection{Metadata}

Participants self-reported the following metadata:
\begin{itemize}[noitemsep,topsep=0pt]
  \item The dialect region that best fits the participant's dialect.
  \item The zip code of the place where the participant grew up or went to school.
  \item Age group (< 19, 19-29, 30-39, 40-49, 50-59, 60-69, 70-79, 80-89, > 89)
  \item Gender (female, male, non-binary)
\end{itemize}

We manually checked the correspondence of reported metadata and recordings for each participant. Collecting the dialect provenance as a zip code allows us to investigate dialects and the performance of speech technologies for them at different granularity levels. Collecting age group and gender helps to make sure that speech technology is inclusive and works across different demographic groups.

\subsection{Recruitment}
For the test set, all participants were recruited via the crowdsourcing platform TestingTime\footnote{\url{https://www.testingtime.com}}. For the train set, half the participants were recruited via TestingTime, whereas the other half were recruited via universities, high schools, newspaper ads, personal contacts, and the crowdsourcing platform seniors@work\footnote{\url{https://www.seniorsatwork.ch}} (for details refer to Appendix~\ref{sec.app.pay} and~\ref{sec.app.consent}).
Only native Swiss German speakers able to correctly translate Standard German to Swiss German were recruited. The goal was to collect the same amount of recordings in each dialect region and we recruited accordingly. The number of recordings per participant was limited to 368 for the test set\footnote{Due to a lack of suitable participants in some dialect regions, 6 participants were allowed to contribute up to 722 recordings.} and 1,112 for the train data. Recruiting the 316 participants required a considerable effort, especially in the low-population regions GR and VS.

\begin{table}[t!]
    \small
    \centering
    \begin{tabular}{l|cccc}
        \textbf{Region} & \textbf{Pop.} & \textbf{Hours} & \textbf{Rec.} & \textbf{Speakers} \\
        \hline
        Basel (BS) & 0.4M & 47.5 & 34,169 & 44 \\
        Bern (BE) & 1.2M & 48.7 & 35,683 & 46 \\
        Grisons (GR) & 0.2M & 44.3 & 30,931 & 46 \\
        Central (CS) & 0.8M & 49.1 & 36,402 & 43 \\
        Eastern (ES) & 0.9M & 52.6 & 38,182 & 47 \\
        Valais (VS) & 0.1M & 51.8 & 36,457 & 44 \\
        Zurich (ZH) & 1.6M & 49.3 & 35,703 & 46 \\
    \end{tabular}
    \caption{Corpus statistics per dialect region. Population is an approximation and only includes German-speaking people .}
    \label{tbl.corpus-reg}
\end{table}

\section{Corpus}
\label{sec.corpus}

The corpus is publicly available\footnote{\url{https://swissnlp.org/datasets/}} under the META-SHARE NonCommercial NoRedistribution license\footnote{\url{http://www.meta-net.eu/meta-share/meta-share-licenses/META-SHARE\%20NonCommercial\%20NoRedistribution-v\%201.0.pdf}}. The distribution format and the included metadata is described in Appendix~\ref{sec.app.distr}. Potential risks are described in Appendix~\ref{sec.app.risks}. The handling of offensive content and personal data is discussed in Appendix~\ref{sec.app.off}.

\begin{figure}[t!]
    \centering
    \includegraphics[width=0.4\textwidth]{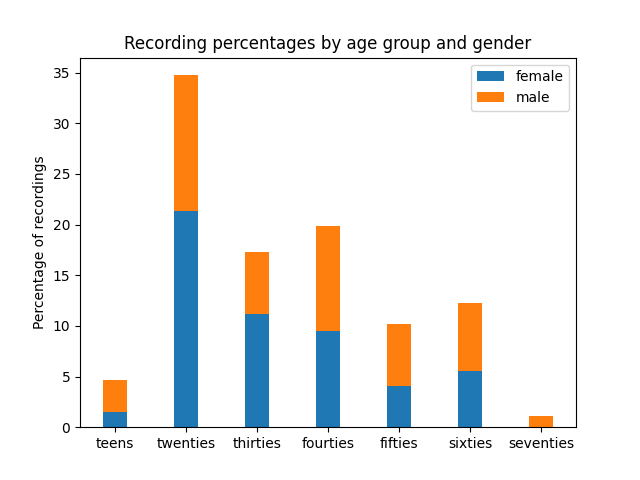}
    \caption{Percentage of recordings by age group and gender}
    \label{fig.age-gender}
\end{figure}

\subsection{Data Cleaning}

\textbf{Filtering.} Recordings with a duration of less than 2 seconds were removed. Silent recordings were also removed. For the test set,  we applied heuristics to flag incomplete sentences, which were removed after double-checking them. We only kept sentences with a recording in all dialect regions in the test set. In total, we filtered out 1.5\% of recordings.
\begin{table}[t]
    \centering
    \small
    \resizebox{.47\textwidth}{!}{%
    \begin{tabular}{l|lll|l}
& train\_all (bal)  & valid & test & full \\ \hline
\textbf{Hours} &  276 (239) &  34  &   34 & 343\\
\textbf{Rec.} &  200K (173K) &   23K  &   25K & 248K\\
\textbf{Unique sent.} &  192K (167K) &   23K  &   4K & 218K\\
\textbf{Speakers} &    219 (192) &  21  &   76 & 316 \\
\textbf{Avg. Rec./speaker} &    $912$ ($902$) &   $1106$  &  $324$ &  $783$\\
\end{tabular}
}
    \caption{Corpus statistics per split. For the train set, the balanced (bal) version is in parentheses.}
    \label{tbl.corpus-spl}
\end{table}
\textbf{Validation.} We validated each speaker manually. For this, we randomly sampled 10 recordings from each speaker, and checked whether the dialect is correct, the recording is in Swiss German, the translation is correct, and whether the sound quality is high enough. All of the participants passed the manual check.

\subsection{Statistics}

The corpus contains 343 hours of Swiss German speech in 247,527 separate recordings, each annotated with the Standard German text translation. The mean recording length is $5.0\pm1.5$ seconds. 217,687 unique sentences were recorded and the vocabulary size is 42,980.
Speech recordings were provided by 316 different speakers, of which 51\% identified as female and 49\% as male. No speaker identified as non-binary. Figure~\ref{fig.age-gender} shows the distribution of the recordings over the age groups, as well as the gender distributions per age group. The age groups from the thirties to the sixties are well represented, while the twenties are overrepresented and the teens as well as seventies are underrepresented. The age groups eighties and above are not represented at all.

Table~\ref{tbl.corpus-reg} shows the corpus statistics per dialect region. While the German-speaking population differs by a factor of up to 16 between regions, the number of recordings per region is a lot more balanced, differing by a factor of not more than 1.2.

\subsection{Splits}
Table~\ref{tbl.corpus-spl} shows the different corpus splits. We provide training, validation, and test splits. There is no speaker overlap between training, validation, and test. There are no common sentences between test and either training or validation. There is, however, an intersection of 835 sentences between training and validation. There are 2 different training splits. train\_all contains all training data, 276 hours of speech. train\_balanced is a subset of train\_all with 239 hours of speech that is balanced in the number of recordings per dialect region. For GR, the region with the fewest recordings, the recordings of all speakers are included in train\_balanced. For the other regions, we randomly chose speakers and added their recordings until the number of GR recordings was reached. train\_balanced includes 33-35 hours of speech, 24,088-25,183 recordings, and 25-32 speakers per region.

Like train\_balanced, the validation split, with 34 hours of speech, is balanced in the number of recordings per dialect region. We randomly chose 3 speakers per region with at least 1,000 recordings. The test set comprises 34 hours of speech. Importantly, the same 3,515 sentences were recorded in all 7 dialect regions to allow a fair comparison between different dialects. The test split contains at least 8 different speakers per region to provide adequate speaker diversity in each region. For this reason, the mean number of recordings per speaker is markedly lower than in the other splits.

\section{Automatic Speech Recognition Baseline}
\label{sec.asr}

\begin{table}
    \small
    \centering
    \resizebox{.5\textwidth}{!}{%
    \begin{tabular}{lll|ll}
        \multirow{2}{*}{\textbf{Dataset}} & \multicolumn{2}{c}{WER} & \multicolumn{2}{c}{BLEU} \\
         & \textbf{validation} & \textbf{test} & \textbf{validation} & \textbf{test} \\ 
        \hline
        ASGDTS & $19.9\pm.1$ & $20.7\pm.3$ & $67.0\pm.2$ & $66.0\pm.4$ \\
        ASGDTS SOTA & $38.7$ & - & $41.9$ & $46.0$ \\ \hline
        SDS-200 & $18.4\pm.1$ & $18.2\pm.1$ & $69.9\pm.1$ & $69.6\pm.1$ \\
        SDS-200 SOTA & $21.7$ & $21.6$ & $63.9$ & $64.0$ \\ \hline
        SPC & - & $30.2\pm.1$ & - & $54.9\pm.2$ \\
        SPC SOTA & - & $23.7$ & - & $60.7$ \\ \hline
        STT4SG-350 & $13.6\pm.1$ & $14.0\pm.1$ & $75.0\pm.1$ & $74.7\pm.1$ \\
    \end{tabular}%
}
    \caption{Performance of the XLS-R Wav2Vec 1B model fine-tuned on the STT4SG-350 train\_balanced split. We report the mean and standard deviation over five different random seeds. ASGDTS: validation = public split, test = private split. We compare each dataset to the state-of-the-art, i.e., ASGDTS SOTA \cite{arabskyy-etal-2021-shared}, SDS-200 SOTA \cite{pluss-etal-2022-sds}, and SPC SOTA \cite{schraner2022swiss}.}
    \label{tbl.model-results}
\end{table}

We train a baseline model to demonstrate the use of the STT4SG-350 corpus for Swiss German ASR. We fine-tune XLS-R (1B)\footnote{Apache-2.0 license} \cite{DBLP:journals/corr/abs-2111-09296} on the train\_balanced split. XLS-R is a model based on wav2vec 2.0 \cite{DBLP:journals/corr/abs-2006-11477} with 965 million parameters pretrained on 436K hours of unlabeled speech data covering more than 128 languages. Swiss German was not part of the training data. We provide the fine-tuning details and experimental setup in appendix~\ref{sec.app.training_details}.

We report the results of our fine-tuned model on three publicly available Swiss German datasets and the STT4SG-350 validation and test sets in Table~\ref{tbl.model-results}.
The model achieves state-of-the-art results on the All Swiss German Dialects Test Set (ASGDTS) \cite{pluss-etal-2021-shared} and SDS-200 \cite{pluss-etal-2022-sds}, and improves the best reported BLEU scores on the test sets by 43\% and 9\%, respectively.
Our model is 6\% behind the best reported BLEU score on the SPC test set \cite{pluss-etal-2021-spc}.
These results highlight the benefit of the STT4SG-350 dataset on test data from different domains.

\section{Conclusion}
\label{sec.concl}

We have described STT4SG-350, which is, to the best of our knowledge, the largest public speech corpus for Swiss German with 343 hours of speech. Our ASR baseline model trained on the corpus achieves a BLEU score of 74.7 on the test set. In addition, it beats the best published BLEU scores on 2 other test sets, demonstrating the quality of the corpus.

STT4SG-350 is balanced across the 7 dialect regions, and the test set allows a fair comparison of ASR performance on different dialects. We intend to take advantage of these properties in future work and conduct in-depth experiments to explore differences in ASR quality between dialects. Subsequently, we want to find ways to improve performance for underrepresented dialects.

\section*{Acknowledgements}
This work was supported by Swiss National Science Foundation within the project "End-to-End Low-Resource Speech Translation for Swiss German Dialects (E2E\_SG)" [205121\_200729/1].

\section*{Limitations}
\label{sec.lim}

The corpus and therefore also the ASR baseline model only cover read speech. We have not tested the model on spontaneous speech, but we expect it to perform significantly worse on this type of data.

Our data collection process for Swiss German speech with Standard German transcripts is designed to collect large amounts of data in a cost-efficient manner. We estimate costs to be 4 to 6 times lower compared to the transcription of existing recordings. However, there is a downside to our approach. Because it is based on a given Standard German sentence, it can lead to Swiss German speech that's closer to Standard German than the Swiss German encountered in everyday conversations. The severity of the shift towards Standard German depends on the individual speakers and their ability and effort to produce Swiss German representations that are close to how they would speak in everyday conversations.

While we made every effort to include as many different dialects as possible in the corpus, there are still strong dialects with a comparatively low German-speaking population that are insufficiently or not at all represented, e.g. some dialects from the canton of Fribourg. This is due to the huge dialect diversity in Switzerland.

The gender ratio is not balanced for some dialect regions in the test set, especially not for VS, where the test set is female-only because we did not succeed to recruit any male speakers from this region during phase 1 of the data collection. However, preliminary experiments do not show a significant difference between genders in Swiss German ASR performance, so we do not expect this to lead to skewed results.

Our ASR baseline model and other models trained on the corpus may perform below average for children and people above seventy due to the lack of training data for these age groups.

\section*{Ethical Considerations}
\label{sec.app.consent}

Participants were specifically recruited to record Swiss German speech for this corpus. The purpose of the recordings was made clear at recruiting time: a training corpus for Swiss German ASR models. Participants were also informed at recruiting time that information about their dialect, age, and gender will be collected. Furthermore, to be able to participate, they had to read and accept our data privacy policy which further detailed the future use of collected data.


\bibliography{custom}
\bibliographystyle{acl_natbib}

\appendix

\section{Web App Screenshot}
\label{sec.app.screen}

\begin{figure*}
\begin{center}
\includegraphics[width=1.0\textwidth]{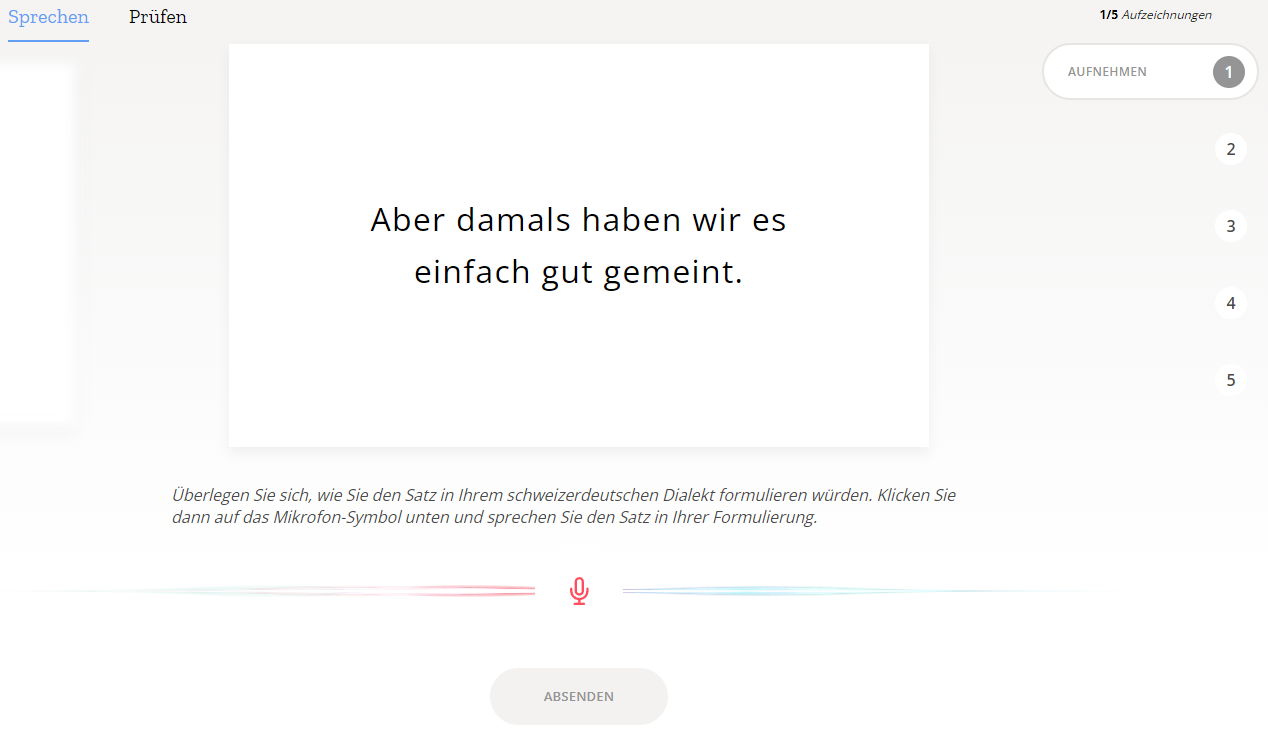}
\caption{Screenshot of the recording functionality in the web app}
\label{fig.app}
\end{center}
\end{figure*}

Figure~\ref{fig.app} shows a screenshot of the recording screen in the web app.

\section{Corpus Distribution Format}
\label{sec.app.distr}

\begin{table*}
    \centering
    \small
    \begin{tabular}{ll}
        \textbf{Column} & \textbf{Description} \\
        \hline
        path & Path to corresponding Swiss German recording in TAR archive \\
        duration & Clip duration in seconds \\
        sentence & Standard German sentence \\
        sentence\_source & Source of the sentence \\
        & news* = Swiss newspapers (for test split: news\_[topic], for other splits: news), \\
        & parliament = parliament minutes of 2 Swiss parliaments \\
        client\_id & Unique speaker identifier \\
        dialect\_region & Speaker's dialect region (Basel, Bern, Graubünden, Innerschweiz, Ostschweiz, Wallis, Zürich) \\
        canton & Canton of the municipality in zipcode column (AG, AI, BE, BL, BS, FR, GL, GR, LU, NW, OW, SG, \\
        & SH, SO, SZ, TG, TI, UR, VS, ZG, ZH, can be empty) \\
        zipcode & Zip code of the origin municipality of a speaker's dialect (can be empty) \\
        age & Speaker's age bracket (teens, twenties, thirties, fourties, fifties, sixties, seventies) \\
        gender & Speaker's gender (female, male) \\
    \end{tabular}
    \caption{Description of columns in TSV files}
    \label{tbl.tsv-cols}
\end{table*}

The recordings are distributed in 2 TAR archives. Recordings in the training and validation splits in FLAC format can be found in clips\_\_train\_valid.tar. Recordings in the test split in MP3 format can be found in clips\_\_test.tar. The mapping of recordings to sentences and all other metadata can be found in the TSV files, one file per split, e.g. train\_all.tsv. A description of the columns in the TSV files can be found in Table~\ref{tbl.tsv-cols}.

\section{Fine-tuning Details}
\label{sec.app.training_details}
The vocabulary used to preprocess the sentences is limited to lower-case characters and the German umlauts ä, ö, and ü. All characters with other accents are transformed into their corresponding character without accents and hyphens are replaced with a space. 

We mainly replicate the fine-tuning procedure\footnote{\url{https://github.com/facebookresearch/fairseq/tree/main/examples/wav2vec/xlsr}} of \citet{DBLP:journals/corr/abs-2111-09296} with the model settings of \citet{DBLP:journals/corr/abs-2006-11477}. Instead of searching the learning rate in a range we settle for $3\mathrm{e}{-5}$. The training is conducted on 4 NVIDIA A100 40 GB GPUs. To achieve an effective batch size of 1,600 seconds (0.44 hours), we use gradient accumulation over 10 steps and 640,000 samples per GPU. One training run on the train\_balance dataset takes 50 hours to complete. The metrics Word Error Rate (WER) and BLEU score are reported as the mean over five runs with different seeds. For the BLEU score, we use the NLTK\footnote{Apache-2.0 license} implementation \cite{BirdKleinLoper09} at version 3.7.

\section{Potential Risks}
\label{sec.app.risks}

The corpus was designed specifically with diversity in mind. The goal was to cover all dialect regions, all age groups and achieve a balanced gender ratio. This goal was reached for the most part. However, no children and people above eighty are part of the corpus. It is possible that models trained on this corpus perform below average for these demographic groups as well as people with strong, not widely used dialects. There is a risk for this group of people to be at a disadvantage when using speech technology solely based on the published corpus.

The described ASR baseline model is intended to be used on Swiss German speech data similar in length to the training data. When transcribing speech that is more than 2 times the mean length of 5 seconds, there is an increasing risk of incomplete transcripts that do not reflect the spoken content well.

\section{Offensive Content and Personal Data}
\label{sec.app.off}

We did not explicitly check for offensive content in the text data because both data sources, newspapers and parliament minutes, are publicly accessible and it seems reasonable to assume that the text does not contain offensive content. This assumption was confirmed by the at least 3,160 recording-sentence-pairs (10 per participant) we manually validated.

We cannot rule out the existence of offensive content in the recordings. However, after the manual validation of at least 3,160 recordings (10 per participant), it is unlikely that there are many such cases.

We did not anonymize data because the metadata doesn't contain information that names or uniquely identifies individuals.

\section{Compensation for Participants}
\label{sec.app.pay}

Participants in the first phase were paid 70 Swiss francs, whereas participants in the second phase were paid 110 Swiss francs. For the last 3 weeks of phase 2, we increased the salary to 200 Swiss francs to attract as many participants as possible before finishing the collection.

Each phase 1 participant should provide 0.5 hours of recordings. Each phase 2 participant should provide 1.5 hours of recordings. We calculated with an hourly salary of 27.50 Swiss francs. 25-30 Swiss francs per hour are the usual payment for a side job in Switzerland. We estimated the required work for each minute of recording to be 2.5 minutes.

For phase 1, the work per participant is therefore 1.25 hours. We added 0.25 hours to read instructions and register on the website. 1.5 times the hourly salary is equal to 41.25 Swiss francs. We increased this to 70 Swiss francs to improve our chances of finding enough participants.

For phase 2, the work per participant is 3.75 hours, plus 0.25 hours setup. 4 times the hourly salary is equal to 110 Swiss francs.

If a participant did not finish the designated amount of recordings, we paid them pro rata.

\end{document}